\documentclass[runningheads]{llncs}
\usepackage[T1]{fontenc}
\usepackage{amssymb}
\usepackage{amsmath}
\usepackage{pifont}
\usepackage{bm}
\usepackage[table]{xcolor}
\usepackage{url}
\usepackage{booktabs} 
\usepackage{multirow} 
\usepackage[colorlinks=true,linkcolor=blue,citecolor=blue,urlcolor=blue,]{hyperref}
\usepackage{algorithm2e}
\usepackage{marvosym}
\usepackage{graphicx}
\begin{document}
\title{Rethinking Cell Counting Methods: Decoupling Counting and Localization}
%
%

\author{Zixuan Zheng\inst{1}\thanks{Equal contribution.} \and
Yilei Shi\inst{1}$^\star$ \and
Chunlei Li\inst{1} \and
Jingliang Hu\inst{1} \and
Xiao Xiang Zhu\inst{2} \and
Lichao Mou\inst{1}\textsuperscript{(\Letter)}}


%
\authorrunning{Z. Zheng et al.}
%
\institute{MedAI Technology (Wuxi) Co. Ltd., Wuxi, China\\\email{lichao.mou@medimagingai.com} \and Technical University of Munich, Munich, Germany}

\maketitle   
\begin{abstract}
Cell counting in microscopy images is vital in medicine and biology but extremely tedious and time-consuming to perform manually. While automated methods have advanced in recent years, state-of-the-art approaches tend to increasingly complex model designs. In this paper, we propose a conceptually simple yet effective decoupled learning scheme for automated cell counting, consisting of separate counter and localizer networks. In contrast to jointly learning counting and density map estimation, we show that decoupling these objectives surprisingly improves results. The counter operates on intermediate feature maps rather than pixel space to leverage global context and produce count estimates, while also generating coarse density maps. The localizer then reconstructs high-resolution density maps that precisely localize individual cells, conditional on the original images and coarse density maps from the counter. Besides, to boost counting accuracy, we further introduce a global message passing module to integrate cross-region patterns. Extensive experiments on four datasets demonstrate that our approach, despite its simplicity, challenges common practice and achieves state-of-the-art performance by significant margins. Our key insight is that decoupled learning alleviates the need to learn counting on high-resolution density maps directly, allowing the model to focus on global features critical for accurate estimates. Code is available at \url{https://github.com/MedAITech/DCL}.

\keywords{cell counting \and decoupling \and message passing }
\end{abstract}

\section{Introduction}
Counting cells in microscopy is of paramount importance in medicine and biology \cite{guo2019sau}. Traditionally, this is done manually, but it is extremely tedious, labor-intensive, and time-consuming. Automated cell counting is therefore vital to many medical and research fields.
\par
In this direction, research has made rapid advances during the past years, driven primarily by the development of counting algorithms in computer vision.
Given an input image, they predominantly learn to predict a full-resolution density map that encodes the quantity and spatial distribution of objects (e.g., people and cells) \cite{liu2019cvpr,cheng2019cvpr}, that is to say, these methods couple counting and localization via the density map. Albeit successful, current state-of-the-art methods tend to increasingly complex designs as diverse as more network modules and additional losses \cite{wang2021miccai,lo2023imai,Ma2022ICIP,Lin2023}. 
\par
In this paper, we break this trend and propose a simple yet effective method for microscopy cell counting. Specifically, we decouple counting and localization, and devise an architecture consisting of two deep networks. The first is a counter that takes an image as input, estimates the number of cells, and generates a coarse, low-resolution density map as a by-product. The second is a localizer that learns to reconstruct a fine-grained localization map conditioned on the original image and the generated coarse density map which implies the estimated cell count. For the counting network, in contrast to prior work performing counting on high-resolution density maps, we propose to do so on intermediate feature representations. Our key insight is that this allows our model to more easily capture global features, without being overwhelmed by fine, low-level details that may not be as relevant. The intermediate representations appear more amenable to learning key attributes needed to produce accurate count estimates.
\par
In addition, since convolution operations in CNNs have limited receptive fields, upon counting touching or overlapping cells, they lack sufficient context to robustly differentiate individual cells. Thus, we propose a global message passing module leveraging patterns and cues across the full image to improve counting accuracy.
\par
We conduct extensive experiments to compare the proposed decoupled learning scheme with other counting methods that jointly learning counting and localization. The competitors include conventional approaches and recent, carefully designed, and more complex models (e.g., using sophisticated network units and losses). From our extensive study across four cell counting datasets, DCC, ADI, MBM, and VGG, we make the following intriguing observations:
\begin{itemize}
    \item We find that decoupling counting and localization has surprising results, which challenges the current common practice in cell counting.
    \item It is advantageous in cell counting to make use of global context. For this reason, we propose a plug-and-play global message passing module.
    \item By applying our decoupled learning scheme, we achieve significantly higher accuracy than well established state-of-the-art methods on multiple cell counting benchmark datasets.
\end{itemize}

\begin{figure}[t]
\centering
\includegraphics[width=\textwidth]{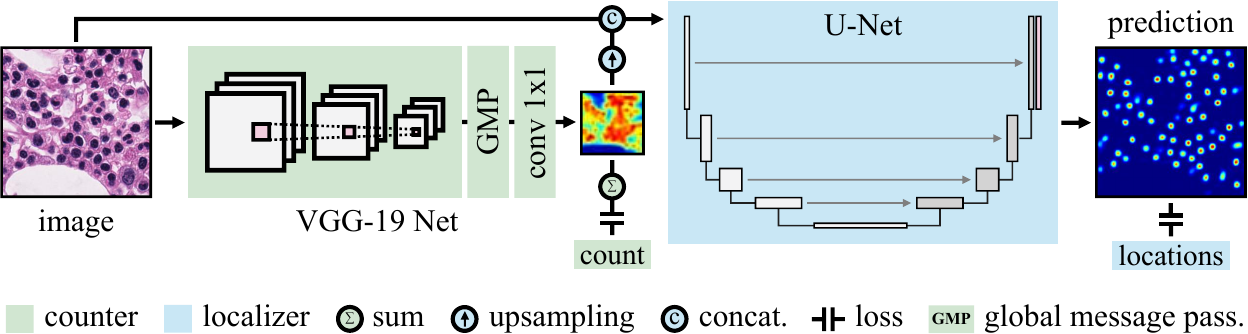}
\caption{The pipeline of the proposed method for cell counting. Unlike prior works, we decouple counting and localization, and design an architecture with separate counter and localizer to enable each to specialize on its target task. Moreover, we introduce a global message passing module into the counting network to model long-range spatial dependencies between image regions, enriching feature representations.} \label{fig_method}
\end{figure}

\section{Methodology}
\subsection{Decoupling Counting and Localization}
Counting and localization are two highly interrelated yet conflicting tasks in cell counting. The former is a more coarse-grained problem that requires richer semantic context, while the latter is more fine-grained and demands detailed information. To address this conflict, we apply decoupled networks. Fig.~\ref{fig_method} shows our method.

\SetKwFunction{flatten}{flatten}

\subsubsection{Counter}
\begin{figure}[t]
\centering
\includegraphics[width=\textwidth]{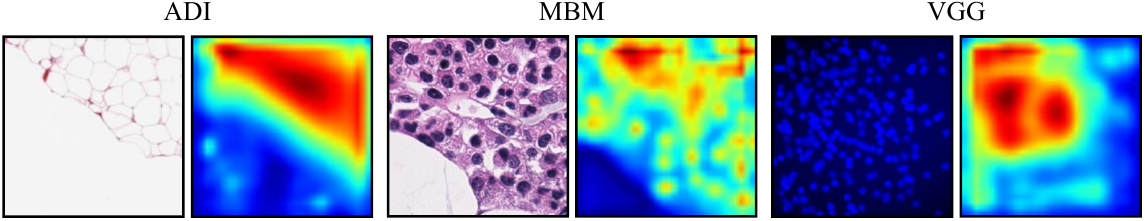}
\caption{Visualization of the learned single-channel feature map in the counting network. Note that the counter manages to localize cells despite having no cell location annotations at training, just cell count labels.} \label{fig_vis_1}
\end{figure}

We use a VGG-19 network \cite{vgg19}, removing its last pooling and fully connected layers, as the backbone for our counting network. Given an image, the output of the backbone is upsampled by a factor of 2 using bilinear interpolation, and then passed to the proposed global message passing module for context modeling. Subsequently, a 1$\times$1 convolutional layer is utilized to generate a single-channel feature map. We make the $\ell_1$ norm of the flattened feature map $\bm{z}$ as close as possible to ground truth count $y$ using the following loss:
\begin{equation}
    \mathcal{L}=|\lVert \bm{z} \rVert_1-y| \,.
\end{equation}
With the designed network architecture and loss, we find that the counter learns a weak, inherent localization ability (see Fig.~\ref{fig_vis_1}). 

\subsubsection{Localizer}
The counting network provides an estimate of the total number of cells, but lacks precise spatial information to localize individual cells. To enable localization, we employ a UNet-based network. By leveraging UNet's capacity for dense prediction, we can reconstruct a fine-grained localization map. Our UNet-based localizer is trained conditioned on both the input image and the predicted coarse density map from the counting network. Learning with this auxiliary guidance allows the localizer to spatially distribute the estimated cell count. We optimize the localization network using mean square error loss. Note that ground truth localization maps are generated via convolving ground truth dot maps with a Gaussian kernel.

\subsection{Global Message Passing}
The proposed global message passing module consists of sampling and feature aggregation, and its goal is to enrich feature representations with long-range dependencies.

\subsubsection{Sampling}
Each feature vector of the convolved image $\bm{x}$ is associated with a spatial index $\bm{p}=(x,y)$. For a query feature vector $\bm{x}_{\bm{p}}$, we consider learning a sampling operating across grids, in order to sample relevant context features to enable longer-range modeling. Specifically, a sampling position $\bm{s}$ can be calculated by applying the following equation:
\begin{equation}
\bm{s}=\bm{p}+\overrightarrow{\omega}(\bm{p})=(x,y)+\overrightarrow{\omega}|_{(x,y)} \,,
\end{equation}
where $\overrightarrow{\omega}=(u,v)$ is a moving vector. Since we would like to learn a task-driven sampling to gather useful context, we define a learnable $\overrightarrow{\omega}$ as follows:
\begin{equation}
\overrightarrow{\omega}=(u(\bm{x}_{\bm{p}}),v(\bm{x}_{\bm{p}})) \,.
\end{equation}
With $u(\cdot)$ and $v(\cdot)$, the sampled position can be predicted, conditioned on the feature of the current position. For the sake of simplicity and more efficient computation, we consider them in the form of linear embeddings, i.e.,
\begin{equation}
u(\bm{x}_{\bm{p}})=\bm{w}_{u}\bm{x}_{\bm{p}} \,,
\end{equation}
\begin{equation}
v(\bm{x}_{\bm{p}})=\bm{w}_{v}\bm{x}_{\bm{p}} \,,
\end{equation}
where $\bm{w}_{u}$ and $\bm{w}_{v}$ are learnable weights and can be implemented as $1\times1$ convolutions.

\subsubsection{Feature Aggregation}
This step aims at aggregating sampled features and generate new feature representations that can facilitate the subsequent cell counting tasks.
\par
We first revisit feature aggregation in self-attention models, which is considered the following equation:
\begin{equation}
\label{eq:attention}
\bm{a}_{\bm{p}}=\sum_{\forall\bm{q}}\frac{1}{\mathcal{C}}w_{\bm{p}\bm{q}}\bm{x}_{\bm{q}} \,.
\end{equation}
Here $\bm{p}$ is a query position whose response $\bm{a}_{\bm{p}}$ is to be calculated, and $\bm{q}$ indicates all possible positions. $w_{\bm{p}\bm{q}}$ represents the relationship between $\bm{p}$ and $\bm{q}$. Moreover, $\mathcal{C}$ is a normalization constant. Eq. (\ref{eq:attention}) is a weighted sum of all feature-map vectors, but learning pairwise relations $\bm{w}=\{w_{\bm{p}\bm{q}}\}$ is computationally expensive.
\par
In order to reduce the computational overhead, in this work, we perform feature aggregation at zero parameters as follows:
\begin{equation}
\label{eq:feat_aggre}
\bm{a}_{\bm{p}}=\sum_{\bm{s}\in\mathcal{V}(\bm{p})}\frac{1}{|\mathcal{V}(\bm{p})|}\bm{x}_{\bm{s}} \,.
\end{equation}
where $\mathcal{V}(\bm{p})$ is a set of sampled positions, conditioned on the position $\bm{p}$. As compared to Eq. (\ref{eq:attention}), our method has two changes: (1) $\forall\bm{q}\rightarrow\bm{s}\in\mathcal{V}(\bm{p})$; (2) $\frac{1}{\mathcal{C}}w_{\bm{p}\bm{q}}\rightarrow\frac{1}{|\mathcal{V}(\bm{p})|}$. By doing so, we achieve feature aggregation in an efficient way.

\section{Experiments}

\subsection{Experimental Setups}

\subsubsection{Datasets}
We evaluate the proposed method on four public benchmarks: the Dublin cell counting (DCC) dataset~\cite{dcc}, the human subcutaneous adipose tissue (ADI) dataset~\cite{adi}, the modified bone marrow (MBM) dataset~\cite{mbm}, and the synthetic fluorescence microscopy (VGG) dataset~\cite{vgg}. To achieve more compelling results given the limited size of these datasets, we partition each into training, test, and validation sets with an approximate ratio of 10:9:1.
\par
The DCC dataset comprises images of various cell types, including embryonic mice stem cells, human lung adenocarcinoma, and human monocytes. Image sizes range from 306×322 to 798×788 to increase diversity.
\par
Sampled from high-resolution histology slides using a $1700\times1700$ sliding window, the ADI dataset contains images with a solution of $150\times150$. Adipocytes within vary dramatically in size and represent a difficult test case given that they are densely packed adjoining cells with few gaps.
\par
The MBM dataset consists of 11 $1200\times1200$ images of bone marrow from 8 healthy individuals, cropped to $600\times600$. The standard staining procedure depicts cell nuclei in blue and other constituents in shades of pink and red.
\par
The VGG dataset is composed of 200 images of size $256\times256$ containing simulated bacterial cells from fluorescence microscopy. The images include overlapping cells at various focal distances, simulating real-life microscopy. Each image contains $174\pm64$ cells.
\par
Table~\ref{tab: datasets} provides an overview of the four datasets and their features.

\begin{table}[!t]
\caption{Overview of four public datasets and their features.}
\label{tab: datasets}
\centering
\renewcommand\arraystretch{1.0}
\setlength{\tabcolsep}{6pt}
\begin{tabular}{rccc}
\hline 
 & Image Size & Cell Count & Type      \\ \hline
DCC                              & $306\times322$ to $798\times788$     & $34\pm22$      & real      \\ 
ADI                              & $150\times150$    & $165\pm44$     & real      \\
MBM                              & $600\times600$    & $126\pm33$     & real      \\
VGG                              & $256\times256$    & $174\pm64$     & synthetic \\
\hline
\end{tabular}
\end{table}

\subsubsection{Evaluation Metrics}
To quantify the cell counting performance of different methods, we employ the widely used mean absolute error (MAE) and mean square error (MSE) metrics~\cite{metric1,metric2}, which measure the discrepancy between the estimated and ground truth cell counts. The MAE is calculated as follows:
\begin{equation}
    \mathrm{MAE}=\frac{1}{N}\sum_{i=1}^{N}|\hat{y}_i-y_i| \,,
\end{equation}
where $\hat{y}_i$ denotes the estimated cell count for the $i$-th test image, and $N$ is the number of test images. MAE is the most commonly used metric in counting tasks. However, a limitation of MAE is its robustness to outliers (i.e., large counting errors). Thus, we additionally report MSE, which is more sensitive to outliers:
\begin{equation}
    \mathrm{MSE}=\frac{1}{N}\sum_{i=1}^{N}(\hat{y}_i-y_i)^2 \,.
\end{equation}

\subsubsection{Implementation Details}
Due to diverse image sizes across datasets, we first preprocess all images as follows. For the DCC dataset, which contains images of different sizes, for each image, we pad the shorter image side to match the scale of the longer side. Images are then resized to the nearest multiple of 256 and divided into non-overlapping $256\times256$ patches organized in a $k\times k$ grid. ADI images are directly resized to $256\times256$. For the MBM dataset, we divide each $1200\times1200$ image into four $300\times300$ patches before resizing to $256\times256$.
\par
For data augmentation during training, we utilize horizontal flipping, vertical flipping, and 90-degree clockwise/counter-clockwise rotations.
\par
We optimize models using Adam with a learning rate of 1e-4 and a batch size of 8. We also employ a cosine learning rate decay scheme with warm restarts~\cite{lr}. Our model is implemented in PyTorch and runs on an NVIDIA RTX 4090 GPU.

\begin{table}[!t]
\caption{Quantitative comparison with state-of-the-art methods on four public datasets. Performance is measured by MAE and MSE.}
\label{tab:cellcout} 
\centering
\renewcommand\arraystretch{1}
\setlength{\tabcolsep}{4pt}
\begin{tabular}{r | cc |cc | cc | cc}
\hline
\multirow{2}{*}{} & \multicolumn{2}{c|}{DCC} & \multicolumn{2}{c|}{ADI} & \multicolumn{2}{c|}{MBM} & \multicolumn{2}{c}{VGG} \\ 
\cline{2-9}
& MAE & MSE & MAE & MSE & MAE & MSE & MAE & MSE \\
\hline
MCNN  & 5.4 & 6.4 & 25.8 & 35.7 & 3.2 & 4.3 & 20.9 & 25.3 \\
FCRN  & 5.6 & 7.3 & 20.6 & 28.3 & 2.8 & 3.7 & 17.7 & 21.5 \\
CSRNet & 2.2 & 2.9 & 13.5 & 18.3 & 2.2 & 2.9 & 7.9  & 10.2 \\
SFCN  & 2.7 & 3.7 & 16.0 & 22.3 & 2.4 & 3.1 & 13.8 & 17.9 \\
DMCount & 2.6 & 3.8 & 9.4 & 13.5 & 2.6 & 3.5 & 6.0 & 8.0 \\
SASNet & 8.9 & 12.0 & 9.0 & 12.2 & 3.9 & 5.2 & 4.9 & 6.8 \\
DQN & 3.5 & 4.6 & 9.7 & 13.1 & 3.1 & 4.2 & 5.5 & 7.3 \\
OrdinalEntropy & 3.2 & 4.3 & 9.1 & 12.0 & 2.9 & 3.8 & 5.7 & 7.8 \\
DiffuseDenoiseCount & 2.8 & 3.7 & 8.8 & 11.9 & 2.9 & 3.9 & 5.5 & 7.0 \\
Ours & \textbf{0.8} & \textbf{1.3} & \textbf{8.4} & \textbf{11.7} & \textbf{1.4} & \textbf{2.1} & \textbf{4.1} & \textbf{5.9}\\
\hline
\rowcolor{gray! 20} improvement & 63.6\% & 55.2\% & 4.5\% & 1.7\% & 36.4\% & 27.6\% & 16.3\% & 13.2\% \\
\hline
\end{tabular}
\end{table}

\subsection{Comparison with State-of-the-Art Approaches}
We compare against state-of-the-art methods including MCNN~\cite{MCNN}, FCRN~\cite{FCRN}, CSRNet~\cite{CSRNet}, SFCN~\cite{SFCN}, DMCount~\cite{DMCount}, SASNet~\cite{SASNet}, DQN~\cite{DQN}, OrdinalEntropy~\cite{OrdinalEntropy}, and DiffuseDenoiseCount~\cite{DiffuseDenoiseCount}. Quantitative results are given in Table~\ref{tab:cellcout}.
\par
On DCC, our model achieves significant improvements over prior arts, reducing MAE and MSE by 63.6\% and 55.2\%, respectively. For ADI, we lower state-of-the-art MAE and MSE by 4.5\% and 1.7\%. On MBM, our method reduces MAE by 36.4\% and MSE by 27.6\% over previous best. Finally, on VGG our approach decreases MAE and MSE by 16.3\% and 13.2\% over state-of-the-art. The consistent advancements across datasets validate the efficacy and generalization of our proposed method.

\begin{table}[t]
\caption{Ablation study results quantifying the impact of the proposed global message passing module on four datasets.}
\label{tab:ablation} 
\centering
\renewcommand\arraystretch{1}
\setlength{\tabcolsep}{3.4pt}
\begin{tabular}{r|cc|cc|cc|cc}
\hline
\multirow{2}{*}{} & \multicolumn{2}{c}{DCC} & \multicolumn{2}{c}{ADI} & \multicolumn{2}{c}{MBM} & \multicolumn{2}{c}{VGG} \\ 
\cline{2-9}
& MAE & MSE & MAE & MSE & MAE & MSE & MAE & MSE \\ 
\hline
w/o GMP & 1.2 & 1.8 & 8.9 & 12.1 & 1.9 & 2.6 & 5.0 & 6.8 \\
Full model & 0.8 & 1.3 & 8.4 & 11.7 & 1.4 & 2.1 & 4.1 & 5.9 \\
\hline
\rowcolor{gray! 20} improvement & 33.3\% & 27.8\% & 5.6\% & 3.3\% & 26.3\% & 19.2\% & 18.0\% & 13.2\% \\ 
\hline
\end{tabular}
\end{table}

\begin{figure}[t]
\centering
\includegraphics[width=\textwidth]{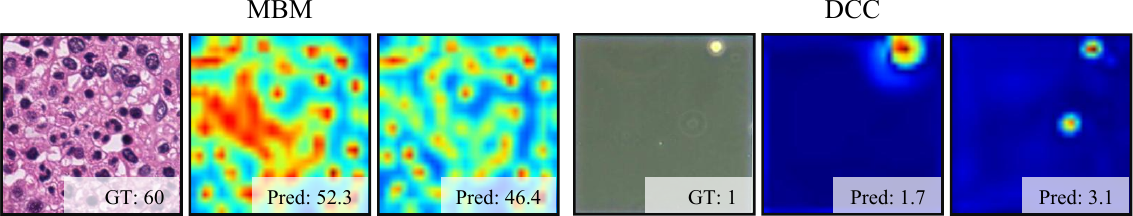}
\caption{Visualization of learned single-channel feature maps from counting networks with and without the proposed global message passing module, along with the estimated cell counts.}
\label{fig_ablation}
\end{figure}

\subsection{Ablation Study} 
We validate the efficacy of our proposed global message passing module via an ablation study, comparing models with and without this component. Removing the module consistently degrades performance across all datasets, as measured by MAE and MSE. For instance, we observe significant drops in MAE and MSE on DCC (33.3\% and 27.8\%) and MBM (26.3\% and 19.2\%) when removing the module. Although smaller, the degradation on ADI (5.6\% MAE, 3.3\% MSE) and VGG (18.0\% MAE, 13.2\% MSE) highlights the module's contribution to precise cell counting. These indicate the module's importance for modeling global context to derive enhanced features. Fig.~\ref{fig_ablation} presents a qualitative ablation study on the effect of this module.

\section{Conclusion}
In this paper, we present a decoupled learning scheme for automated cell counting in microscopy images. Our approach deviates from the common practice of jointly learning counting and density map estimation, instead decoupling these objectives into separate counter and localizer networks. The counter operates on intermediate feature maps to leverage global context and produce coarse density maps along with count estimates, while the localizer reconstructs high-resolution density maps to localize individual cells, conditioned on the input images and coarse density maps. To further enhance counting accuracy, we introduced a global message passing module to integrate cross-region patterns.
\par
Through extensive experiments on four datasets, we demonstrate that our decoupled learning approach challenges the common practice and achieves state-of-the-art performance by significant margins, despite its simplicity.

\begin{credits}
\subsubsection{\discintname}
The authors have no competing interests to declare that are relevant to the content of this paper.
\end{credits}

%
%
%
%

\end{document}